%% file: iclr2027_conference.tex
\documentclass{article} 
\usepackage{iclr2027_conference,times}
\usepackage[pdftex]{graphicx} 

\input{math_commands.tex}

\usepackage{hyperref}
\usepackage{url}
\newcommand{\sable}{\textsc{sable}}

\title{Sparse-View Interpretable 3D Animal Behavior Representations for Neural Encoding and Decoding}

\author{\textbf{Xinming Dai$^{*1}$ \quad Qihang Jin$^{*2}$ \quad Jiaying Zhang$^{*1}$ \quad Tianshu Tan$^3$ \quad Baiyuan Chen$^4$} \\
\textbf{Hanrui Lyu$^5$ \quad Lenny Aharon$^1$ \quad Kyle Daruwalla$^6$ \quad Xun Helen Hou$^6$} \\
\textbf{Matthew R. Whiteway$^1$ \quad Liam Paninski$^{\dagger,1}$ \quad Yizi Zhang$^{\dagger,1}$} \\
\AND
\textmd{$^1$Columbia University \quad $^2$University of Science and Technology of China} \\
$^3$Harvard University \quad $^4$University of Cambridge \\
$^5$Northwestern University \quad $^6$Cold Spring Harbor Laboratory \\
$^*$Equal contribution \\
$^{\dagger}$Equal advising \\
Correspondence to: xd2319@columbia.edu, qihangjin@mail.ustc.edu.cn
}

\iclrfinalcopy 
\begin{document}

\maketitle

\begin{abstract}
A deeper understanding of brain function requires a precise, structured characterization of behavior. Yet, extracting behavioral representations from video in a form suitable for scientific analysis remains a fundamental challenge. Many prior studies represent behavior via pose estimation or nonlinear video embeddings. However, pose tracking discards rich information beyond predefined keypoints, while nonlinear video embeddings lack interpretability. We address this limitation with {\sable} (\textbf{S}parse-view \textbf{A}nimal \textbf{B}ehavior \textbf{L}atent \textbf{E}mbeddings), a self-supervised framework that leverages a geometric inductive bias to learn behavior representations. By augmenting a multi-view transformer with priors from monocular depth and pose estimation, {\sable} reconstructs 3D animal behavior from extremely sparse views while learning explicit 3D latent structure. Without ground-truth 3D labels, it reliably recovers 3D behavior from two-view videos, whereas state-of-the-art (SOTA) methods fail or yield degenerate solutions. Across the International Brain Lab and Cheese3D datasets, we demonstrate that {\sable} learns 3D representations that match or exceed prior SOTA performance in neural encoding and decoding. Single-neuron analyses further identify neurons with putative 3D tuning, generating testable hypotheses about how the brain represents 3D behavior. Once pretrained across animals, {\sable} serves as an off-the-shelf model that generalizes zero-shot to unseen animals without animal-specific calibration or retraining. Our method establishes 3D-aware video embeddings that capture complex behavior, opening new avenues for studying brain–behavior relationships.
\end{abstract}

\begin{figure}[ht]
    \centering
    \includegraphics[width=\textwidth]{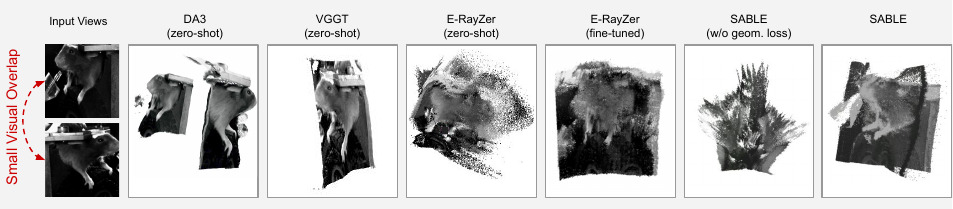}
    \caption{\small{\bf Geometric inductive bias enables {\sable} to reliably reconstruct 3D animal behavior from sparse-view IBL data, where existing 3D foundation models fail to recover coherent 3D structure.} Existing approaches struggle to recover coherent 3D structure in this challenging two-camera setting. Depth Anything 3 (DA3) \citep{lin2025depth} produces plausible monocular point clouds from depth estimation but fails to infer camera poses for cross-view fusion. Visual Geometry Grounded Transformer (VGGT) \citep{wang2025vggt}, applied zero-shot, collapses to a degenerate single-view 2D shortcut solution, and E-RayZer \citep{zhao2026erayzer} exhibits a similar failure mode; even after finetuning, it produces overlapping, flattened point clouds. In contrast, {\sable} learns accurate camera parameters, avoids degenerate shortcut solutions, and reconstructs coherent 3D mouse structure by incorporating geometric inductive bias. Without the corresponding geometric loss (“NoGeom”), reconstruction degenerates.
    \label{fig:3d_recon_ibl}}
\end{figure}

\section{Introduction}

Neural activity reflects and shapes interactions with the external environment, making behavior a primary lens for understanding the brain \citep{krakauer2017neuroscience, datta2019computational}. Advances in large-scale video recordings enable detailed behavioral measurements \citep{gomez2014big, berman2018measuring}, motivating methods that extract informative features from high-dimensional video \citep{pereira2020quantifying}. Pose estimation methods \citep{mathis2018deeplabcut, pereira2019fast, biderman2024lightning, aharon2026lightning} provide interpretable summaries of behavior but miss its full complexity, while nonlinear video embeddings \citep{batty2019behavenet, wang2026animal} capture richer information but can be challenging to interpret~\citep{whiteway2021partitioning}.
Moreover, both are typically computed from 2D projections of an animal that moves in three dimensions. Capturing behavior in 3D is essential for resolving viewpoint ambiguities and recovering fine-grained movements that cannot be inferred from 2D observations alone \citep{marshall2022leaving}. What is needed, then, are methods that extract dense, interpretable 3D representations of behavior from video.

Recent computer vision foundation models have demonstrated strong 3D reconstruction performance from multi-view images and videos \citep{lin2025depth, chen2025videodepthanything, wang2025vggt, zhao2026erayzer}. However, applying these approaches to animal behavioral data remains challenging, particularly in neuroscience where video distributions differ from standard benchmarks and camera views are sparse with minimal visual overlap \citep{yao2022lassie}. In particular, two-view imaging is a common configuration in animal behavior experiments \citep{musallSingletrialNeuralDynamics2019, warrenRapidWhiskerbasedDecision2021}, providing the minimal setup to recover 3D animal kinematics through stereo reconstruction \citep{liCorticothalamicCommunicationAction2024,grierMouseSensorimotorCortex2026}. Compared with larger multi-camera systems, a two-camera setup is more affordable, compact, and easier to deploy. Yet, current 3D foundation models are trained on datasets with dense visual overlap---often 10 views or more \citep{wang2025vggt, zhao2026erayzer}---and therefore struggle with the limited visual overlap present in the two-view setting. A representative example is the two-view dataset from the International Brain Laboratory (IBL) \citep{international2025brain}, which captures a mouse turning a wheel in a visual decision-making task. On this dataset, existing 3D foundation models often produce degenerate reconstructions that fail to capture the animal’s behavior (Fig.~\ref{fig:3d_recon_ibl}). These limitations motivate models that can recover detailed 3D animal geometry from the \textit{sparse} and \textit{minimally overlapping} views typical of neuroscience experiments.

We address these challenges with a self-supervised 3D reconstruction framework that leverages inductive biases to learn explicit 3D behavioral representations from sparse views for downstream neural encoding and decoding. {\sable} (\textbf{S}parse-view \textbf{A}nimal \textbf{B}ehavior \textbf{L}atent \textbf{E}mbeddings) employs masked modeling to reconstruct 3D Gaussian point clouds from multi-view video. Built on a multi-view transformer pretrained on diverse datasets with overlapping views \citep{zhao2026erayzer}, it is fine-tuned on sparse, minimally overlapping two-view videos. By incorporating pseudo-point clouds derived from monocular depth and animal pose keypoints as geometric priors, {\sable} regularizes the learned 3D structure to avoid degenerate solutions. We demonstrate that {\sable} reconstructs high-fidelity 3D point clouds from IBL videos using only two sparse views and no 3D labels, whereas existing 3D models fail in this regime (Fig.~\ref{fig:3d_recon_ibl}). We further apply {\sable} to Cheese3D \citep{daruwalla2026cheese3d}, a head-fixed mouse dataset comprising neural recordings and behavioral videos from six camera views, demonstrating its generalization beyond the two-view setting. Through comprehensive neural encoding on both datasets and neural decoding on IBL, we show that {\sable} matches or outperforms previous SOTA in predicting neural activity and reconstructing video frames. Moreover, {\sable} enables consistent decoding of 3D animal behavior directly from neural recordings and, once pretrained, generalizes zero-shot to unseen animals. To the best of our knowledge, this is the first demonstration that detailed 3D animal behavior can be reliably decoded from intracortical neural activity. The contributions of this work include:
\begin{itemize}
    \item A self-supervised framework for reconstructing interpretable, stable 3D animal behavior from extremely sparse multi-view video using explicit geometric inductive biases, overcoming the degenerate solutions of prior 3D models.
    \item A new paradigm for neural encoding and decoding that leverages fine-grained 3D behavioral structure to better capture brain--behavior relationships.
    \item A pretrained model that generalizes to unseen animals with minimal adaptation and is readily deployable in labs collecting head-fixed mouse neural recordings and behavioral videos.
\end{itemize}

\section{Related Work}

\paragraph{Sparse keypoint representations for 3D animal behavior analysis.} Deep learning has  advanced video-based animal behavior quantification \citep{mathis2020deep, pereira2020quantifying}, commonly through per-view 2D keypoint estimation \citep{mathis2018deeplabcut,pereira2022sleap,biderman2024lightning} followed by 3D triangulation \citep{nath2019using, karashchuk2021anipose}. Recent methods demonstrate that this pipeline can be improved by integrating information across views before the final 3D prediction: DANNCE fuses multi-view image features into a volumetric representation \citep{dunn2021geometric}, while Lightning Pose 3D uses a multi-view transformer with 3D reprojection supervision \citep{aharon2026lightning}. However, these supervised methods produce only sparse keypoints, rather than dense animal geometry or appearance.

\paragraph{Large-scale foundation models for 3D reconstruction.}
Recent 3D foundation models mark a paradigm shift toward direct, feed-forward inference of 3D structure from multi-view data \citep{wang2024vggsfm, wang2025vggt}. Visual Geometry Grounded Transformer (VGGT) jointly estimates cameras, depth, and point clouds with geometric supervision \citep{wang2025vggt}. Depth Anything 3 (DA3) predicts consistent depth and ray maps \citep{lin2025depth}; and E-RayZer learns 3D-aware representations self-supervised through Gaussian-splatting reconstruction \citep{zhao2026erayzer}. Despite this progress, sparse, minimally overlapping animal views are out of distribution for these models, often causing unstable camera poses, degenerate geometry, or 2D collapse (Fig.~\ref{fig:3d_recon_ibl}). This motivates domain-adapted geometric priors for 3D animal behavior reconstruction.

\paragraph{Learning behavior representations for neural analysis.}
Existing approaches learn behavioral representations for downstream neuro-behavioral analyses, but they remain limited to 2D or depend on dense multi-view setups. For example, BEAST learns video representations for neural encoding \citep{wang2026animal}, but its ImageNet-pretrained ViT backbone \citep{he2022masked} primarily captures 2D semantic information while overlooking cues encoded in structure, motion, and depth. MoReMouse reconstructs dense 3D mouse pose using a canonical mouse avatar learned from synthetic data \citep{zhong2026moremouse}, but this reliance on synthetic data limits its applicability to real-world sparse-view recordings. Pose Splatter \citep{goffinet2025pose} and BEAST3D \citep{wang2026beast3danimalbehavioralanalysis} were both developed for animal behavior analysis and learn dense 3D representations, but both require at least three to four camera views and cannot operate in the two-view setting. Thus, existing approaches cannot learn dense 3D behavioral representations from sparse views.

\section{Dataset}\label{sec:data}
The IBL Brainwide Map dataset records brain-wide neural activity and behavior in head-fixed mice performing a standardized visual decision task: mice discriminate visual stimuli presented on the left or right side of a screen and report their choice by turning a response wheel~\citep{international2025brain}. To show generalizability, we also evaluate {\sable} on Cheese3D \citep{daruwalla2026cheese3d}, an additional head-fixed mouse dataset (Appendix~\ref{app:cheese3d_data}).

\paragraph{Neural data.} Neural activity is measured using Neuropixels probes, which provide high-density, multichannel recordings of spiking activity from hundreds of neurons simultaneously \citep{jun2017fully}. In our experiments, we analyze data from ten sessions across ten animals, using seven for pretraining and holding out three to test generalization to unseen data. Each recording session includes up to two probes, with insertion sites varying across animals to sample diverse brain areas. From each held-out session, we randomly sample neural activity from 400 trials, yielding 1,200 trials in total, each lasting 1 second. Spike trains are binned at $\sim$16.7 ms resolution (60 bins per 1s trial, matching the 60 Hz video frame rate), and neurons with mean firing rates below 5 Hz are excluded. In total, we analyze 2,619 units from 6 probe insertions across 3 mice, covering 37 anatomical regions.

\paragraph{Behavioral data.} Behavioral measurements include wheel movements and video-based pose estimates \citep{biderman2024lightning} from the held-out sessions (Appendix~\ref{app:lp_pose_ibl}). In each session, neural recordings are paired with synchronized videos from two cameras (left and right) capturing wheel-turning behavior. The cameras record at 60 Hz and 150 Hz; the latter is resampled to 60 Hz ($\sim$16.7 ms per frame) via nearest-neighbor interpolation. The two views have limited overlap, primarily covering the nose, forepaws, and wheel (see Fig. \ref{fig:3d_recon_ibl}). In each session used for pretraining, we select 1,000 frames using a Principal Component Analysis (PCA) + $k$-means strategy to capture diverse behaviors (Appendix~\ref{app:frame_selection_ibl}) for model training. For evaluation, we extract frames aligned with the timestamps of the 400 sampled neural trials in each session, yielding a total of 24,000 pairs of frames per session for downstream analyses.

\section{Methods}

\begin{figure}[!ht]
    \centering
    \includegraphics[width=\textwidth]{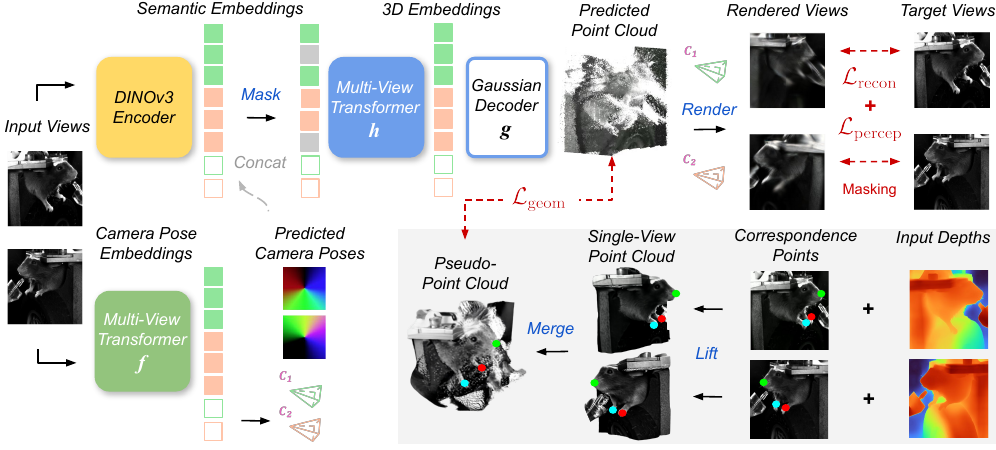}
    \caption{\small{\bf {\sable} framework.} A multi-view transformer predicts camera poses from two input views, which are converted into Plücker ray maps. These rays are concatenated with image tokens from a DINOv3 encoder that capture semantic information, masked with learnable mask tokens, and processed by a second multi-view transformer to produce 3D-structured embeddings that parameterize a Gaussian splatting decoder for 3D reconstruction. To prevent shortcut solutions, we impose geometric regularization via pseudo-point clouds derived from monocular depth and aligned across views using animal pose keypoints as correspondences. The predicted Gaussian point cloud is projected using the estimated camera poses and rendered into 2D images. Training is guided by a masked view reconstruction loss $\mathcal{L}_{\text{recon}}$ for pixel-level accuracy, a perceptual loss $\mathcal{L}_{\text{percep}}$ for perceptual similarity, and a geometric regularization term $\mathcal{L}_{\text{geom}}$ that constrains the 3D locations of the predicted point clouds. \label{fig:model}}
\end{figure}

\paragraph{Our insights.} 
Self-supervised 3D learning from sparse, minimally overlapping views is ill-posed: reconstruction alone often collapses to trivial 2D solutions \citep{younis2025sparse}. We address this by supervising predicted 3D point clouds with pseudo-point clouds derived from external depth and pose models. The distilled geometry provides an inductive bias that constrains the solution space and discourages trivial shortcuts and favors geometry-consistent representations for downstream neural analyses.

\paragraph{Overview.} As shown in Fig. \ref{fig:model}, {\sable} predicts camera parameters with a multi-view transformer, then combines masked DINOv3 image tokens \citep{siméoni2025dinov3} with ray tokens from a second multi-view transformer to learn 3D-structured embeddings. These embeddings parameterize a Gaussian-splatting decoder that reconstructs 3D point clouds. Both transformers are initialized from E-RayZer \citep{zhao2026erayzer} to inherit multi-view and camera-pose priors. To prevent shortcut solutions, we regularize predictions using pseudo-point clouds derived from monocular depth, which are aligned across views using Lightning Pose keypoints \citep{aharon2026lightning} as correspondences. A differentiable renderer projects the point clouds back into 2D views. After multi-animal pretraining, {\sable} generalizes zero-shot to unseen animals without dataset-specific adaptation. 

\paragraph{Notation.}
For each video frame, we consider $V$ views of input images $I = \{I^{(v)}\}_{v=1}^V$, each with resolution $H \times W$ and $C$ channels, and tokenize each view into $h \times w$ patches, where $h = H/p$ and $w = W/p$ denote the number of patches along the height and width, respectively, for patch size $p$. We denote tokens by $\mathbf{t}$, intermediate features by $\mathbf{e}$, and learned embeddings by $\mathbf{z}$. 
We further denote depth maps by $\mathbf{d}$ and point clouds by $\mathbf{p}$, which represent the 3D positions of points.

\subsection{Model Architecture}\label{sec:model_architecture}

\paragraph{Multi-view transformer.}
{\sable} comprises two E-RayZer–style multi-view transformers with alternating within-view and cross-view attention~\citep{zhao2026erayzer}. We tokenize all input views into patches to obtain image tokens 
$\mathbf{t}_{\text{view}} \in \mathbb{R}^{(Vhw) \times D}$, with $D$ being the embedding dimension. A first transformer $f$ encodes these tokens along with prepended camera pose tokens $\mathbf{t}_{\text{cam}}$ into intermediate features
\begin{equation}
\mathbf{e}_{\text{cam}}, \mathbf{e}_{\text{view}} = f(\mathbf{t}_{\text{cam}}, \mathbf{t}_{\text{view}}), \quad \mathbf{e}_{\text{cam}} \in \mathbb{R}^{V \times D}, \, \mathbf{e}_{\text{view}} \in \mathbb{R}^{(Vhw) \times D}.
\end{equation}
$\mathbf{e}_{\text{cam}}$ are used to predict camera intrinsics and extrinsics, which are converted into Plücker ray tokens 
$\mathbf{r}_{\text{plk}} \in \mathbb{R}^{(Vhw) \times D}$ \citep{hartley2003multiple}. 
We further process the image tokens using a DINOv3 encoder with a ViT-B backbone \citep{dosovitskiy2021image}, and then concatenate and fuse the image and ray tokens using a multi-layer perceptron (MLP) $l$. Per-view summary [CLS] tokens $\mathbf{e}_{\text{view}}^{\text{[CLS]}} \in \mathbb{R}^{V \times D}$, inherited from the DINOv3 [CLS] tokens, are also prepended. This sequence of tokens is masked with learnable mask tokens and passed to a second transformer $h$ to produce 3D-aware embeddings:
\begin{equation}
(\mathbf{z}_{\text{view}}^{\text{[CLS]}}, \mathbf{z}_{\text{view}})
= h\big(\mathbf{e}_{\text{view}}^{\text{[CLS]}},\, l(\mathbf{e}_{\text{view}}, \mathbf{r}_{\text{plk}})\big). \label{eq:view_embed}
\end{equation}
Unlike E-RayZer \citep{zhao2026erayzer}, which masks entire views (often many at once) and reconstructs them from the remaining ones, {\sable} is tailored to the extreme two-view setting. In this regime, removing one view leaves only a single view, providing insufficient geometric constraints for 3D recovery and leading to 2D collapse. Instead, {\sable} applies patch-level masking across both views to preserve cross-view context and reduce visual ambiguity.

\paragraph{Gaussian splatting decoder and renderer.}
Given per-token image patch embeddings $\mathbf{z}_{\text{view}}$, we use a lightweight linear decoder $g$ to predict per-pixel Gaussian parameters via upsampling:
\begin{equation}
\mathcal{G} = g(\mathbf{z}_{\text{view}}) \in \mathbb{R}^{(VHW) \times F},
\end{equation}
where $F$ denotes the number of Gaussian parameters. Each element of $\mathcal{G}$ corresponds to a pixel-aligned 3D Gaussian primitive. 
For each Gaussian $i$, the parameters include 3D position $\mathbf{p}_i \in \mathbb{R}^3$, orientation represented as a quaternion $\mathbf{q}_i \in \mathbb{R}^4$, scale $\mathbf{s}_i \in \mathbb{R}^3$, opacity $\alpha_i \in \mathbb{R}$, and spherical harmonic coefficients $\mathbf{C}_i \in \mathbb{R}^{(d_{SH}+1)^2 \times 3}$~\citep{kerbl20233d}. 
The predicted Gaussians are projected into each view using the camera intrinsics and extrinsics estimated by transformer $f$, and rendered into reconstructed 2D images $\hat{I}$ via a differentiable Gaussian splatting renderer.

\paragraph{Pseudo-point cloud generation.}
We obtain monocular depth maps $\mathbf{d}^{(v)} \in \mathbb{R}^{HW}$ for each view $v \in \{1,\dots,V\}$ using Video Depth Anything (VDA)~\citep{chen2025videodepthanything}, and normalize them to  $\tilde{\mathbf{d}}^{(v)} \in [-1,1]^{HW}$. We further define normalized image-plane coordinates $(\tilde{x}^{(v)}_i, \tilde{y}^{(v)}_i) \in [-1,1]^2$ for each pixel $i$. Each pixel is lifted to a 3D point in a pseudo-coordinate space $\tilde{\mathbf{p}}^{(v)}_i = (\tilde{x}^{(v)}_i, \tilde{y}^{(v)}_i, \tilde{d}^{(v)}_i) \in \mathbb{R}^3$.
Aggregating all pixels yields a per-view pseudo-point cloud 
$\tilde{\mathbf{p}}^{(v)} \in \mathbb{R}^{HW \times 3}$. 
For two views, we denote the corresponding pseudo-point clouds as 
$\tilde{\mathbf{p}}^{(\text{src})}, \tilde{\mathbf{p}}^{(\text{tgt})} \in \mathbb{R}^{HW \times 3}$. 

We use a pose tracking algorithm~\citep{aharon2026lightning} to identify $N \ll HW$ keypoints in each 2D view (e.g., nose, left paw, and right paw), and treat their lifted 3D locations in the pseudo-coordinate space as correspondences for view merging. Let 
$\{\tilde{\mathbf{u}}^{(\text{src})}_n\}_{n=1}^N$ and $\{\tilde{\mathbf{u}}^{(\text{tgt})}_n\}_{n=1}^N$ denote these corresponding points from both pseudo-point clouds. For each frame, we then estimate a rigid transformation via the Kabsch algorithm~\citep{kabsch1976solution}:
\begin{equation}
\min_{\mathbf{R} \in SO(3),\, \mathbf{T}} 
\sum_{n} \|\mathbf{R}\tilde{\mathbf{u}}^{\text{(src)}}_n + \mathbf{T} - \tilde{\mathbf{u}}^{\text{(tgt)}}_n\|_2^2.
\end{equation}
The optimal rotation $\mathbf{R}$ and translation $\mathbf{T}$ are applied to map the source point cloud to the target coordinate space $\tilde{\mathbf{p}}^{\text{(src)}'} = \mathbf{R}\tilde{\mathbf{p}}^{\text{(src)}} + \mathbf{T}$. Finally, we concatenate the source and target point clouds represented in the same coordinate system into a single set to obtain a pseudo-point cloud for geometric regularization.


\subsection{Training Objectives}

\paragraph{Masked reconstruction and perceptual losses.}
For each frame, let $\hat{I}$ and $I$ denote the rendered and target images and $\mathcal{M}$ the set of masked pixels across all views. We compute:
\begin{equation}
\mathcal{L}_{\text{recon}}  
+ \mathcal{L}_{\text{percep}}
= \frac{1}{|\mathcal{M}|} \sum_{i \in \mathcal{M}} \|\hat{I}_i - I_i\|_2^2
+ \sum_{l} \frac{1}{|\phi_l|} \|\phi_l(\hat{I}) - \phi_l(I)\|_1,
\end{equation}
where $\phi_l(\cdot)$ denotes features from layer $l$ of a pretrained VGG-19 network \citep{simonyan2015very}. 

\paragraph{Geometric regularization.}
Let $\hat{\mathbf{p}}, \tilde{\mathbf{p}} \in \mathbb{R}^{(VHW) \times 3}$ denote the predicted and pseudo point clouds for each frame. 
We first normalize both point clouds to remove global translation and scale differences before enforcing geometric consistency via:
\begin{equation}
\mathcal{L}_{\text{geom}} = \frac{1}{VHW}\sum_{i=1}^{VHW} \|\hat{\mathbf{p}}_i - \tilde{\mathbf{p}}_i\|_2^2. \label{eq:geom_loss}
\end{equation}
The final objective is a weighted sum of losses, with $\lambda$ denoting the weights (Appendix~\ref{app:model_hyperparams}):
\begin{equation}
\mathcal{L} 
= \lambda_{\text{recon}} \cdot \mathcal{L}_{\text{recon}} 
+ \lambda_{\text{percep}}  \cdot \mathcal{L}_{\text{percep}} 
+ \lambda_{\text{geom}} \cdot \mathcal{L}_{\text{geom}}.
\label{eq:model_formula}
\end{equation}

\section{Results}

\subsection{3D Animal Behavior Reconstruction}\label{sec:recon}
Learning 3D animal pose is essential for capturing body–environment interactions beyond ambiguous 2D projections and for studying their neural basis. On IBL data, we show that our method reconstructs accurate animal point clouds from only two minimally overlapping views, whereas state-of-the-art 3D models collapse to degenerate solutions. We further validate this on Cheese3D \citep{daruwalla2026cheese3d}; see Appendix~\ref{app:nvs_cheese3d} for results.

\paragraph{Models.} We benchmark point cloud reconstruction quality in sparse-view settings against the following large-scale 3D models: (1) \textit{VGGT} employs a multi-view transformer with alternating local–global attention to directly infer 3D attributes of a scene from multiple views and is trained with 3D supervision \citep{wang2025vggt}. (2) \textit{DA3} uses a DINO-style transformer with cross-view attention to predict depth via teacher–student training and ray maps using a dual-DPT head; point clouds are then obtained by back-projecting pixels along the predicted rays using the estimated depth values \citep{lin2025depth}. (3) \textit{E-RayZer} is a self-supervised model that adopts VGGT-style multi-view transformers for camera pose estimation and Gaussian splatting for cross-view reconstruction \citep{zhao2026erayzer}.

\paragraph{Evaluation.}
We qualitatively compare predicted 3D point clouds across baseline models on IBL data captured from two camera views. Since ground-truth 3D labels are unavailable, models that rely on 3D supervision, such as VGGT, are evaluated only in a zero-shot setting using pretrained checkpoints. We also report zero-shot results for \textit{DA3} and E-RayZer. Because E-RayZer is self-supervised and does not require 3D labels, we additionally finetune it on the IBL two-camera dataset and visualize the resulting point clouds. {\sable} is trained under the same setting as E-RayZer for a direct comparison.

\paragraph{Results.} As shown in Fig.~\ref{fig:3d_recon_ibl}, 3D point cloud reconstruction from IBL data is particularly challenging due to minimal cross-view overlap (e.g., nose, paws, and wheel), distribution shift from standard 3D datasets, and the absence of ground-truth 3D supervision. Under these conditions, existing large-scale 3D models fail to recover coherent geometry. DA3 leverages monocular depth cues to produce plausible single-view point clouds, but fails to estimate accurate camera poses, so that the final point cloud is misaligned across views. VGGT collapses to a degenerate single-view 2D solution, and E-RayZer, when applied zero-shot, exhibits a similar failure mode; even after finetuning, E-RayZer produces overlapping, flattened point clouds due to inaccurate camera pose estimation. In contrast, {\sable} learns consistent camera parameters and reconstructs coherent 3D mouse structure using a geometry loss based on pseudo-point clouds, which provides a geometric inductive bias. Without this loss, the reconstruction degenerates. These interpretable 3D representations support downstream neural analyses, as detailed in the following sections.

\begin{figure}[t!]
    \centering
    \includegraphics[width=\textwidth]{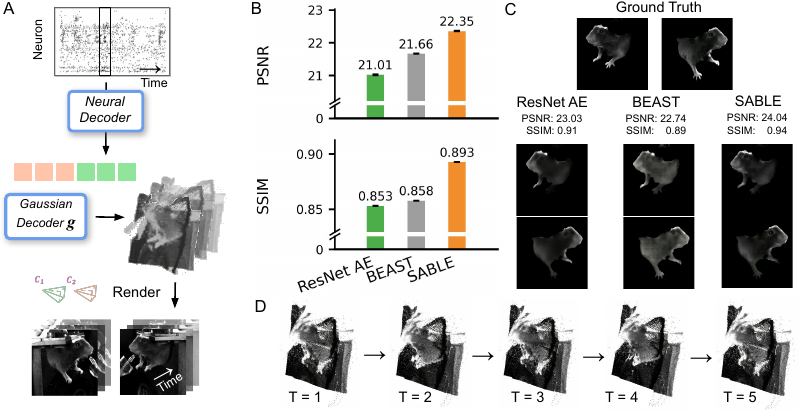}
    \caption{\small{\bf 3D behavioral dynamics can be accurately decoded from neural activity.} (A) For each video frame, we train a TCN decoder to predict image patch embeddings from neural activity, which are then passed to a Gaussian decoder to generate 3D point clouds and render 2D images. (B) Decoding performance is evaluated across baseline models that learn video-derived behavioral embeddings. ResNet AE compresses image features extracted with a ResNet-18 backbone into latent vectors, while BEAST and {\sable} use image patch embeddings produced by a ViT-MAE and a multi-view transformer, respectively. {\sable} outperforms all baselines in video frame decoding quality, as measured by PSNR and SSIM, which evaluate pixel-level accuracy and perceptual quality. Error bars indicate the standard error of the mean across 14,400 frame pairs from held-out sessions. (C) Comparison of decoded video frames from ResNet AE, BEAST, and {\sable} against the ground truth. (D) The Gaussian point clouds decoded from neural activity capture fine-grained paw movements. \label{fig:decoding}}
   
\end{figure}

\subsection{Neural Decoding}\label{sec:decoding}
Neural decoding measures how well neural activity predicts behavior and underlies brain–computer interfaces such as motor and speech prosthetics \citep{silvaSpeechNeuroprosthesis2024}. Here, we show that our method learns behavioral representations that are more accurately decoded from neural activity, enabling reconstruction of 3D animal behavior. We focus on IBL, which provides a stronger decoding signal with at least 708 neurons per session, versus 8 in Cheese3D.

\paragraph{Models.} We benchmark the following nonlinear baselines on decoding video frames from neural activity: (1) \textit{ResNet autoencoder (AE)}: We use a ResNet-18 backbone to extract frame features \citep{he2016deep}, compress them into nonlinear latents via an MLP encoder, and reconstruct images with an MLP decoder and another ResNet-18~\citep{batty2019behavenet}. Video embeddings are constructed by concatenating latents from both views. (2) \textit{BEAST}: We finetune an ImageNet-pretrained ViT-MAE \citep{he2022masked} with a ViT-B backbone \citep{dosovitskiy2021image} on IBL data to learn per-view image patch embeddings, which are fed into a ViT-MAE decoder for frame reconstruction \citep{wang2026animal}. (3) {\sable}: The multi-view transformers $f$ and $h$ are initialized from a pretrained E-RayZer model (Appendix \ref{app:model_details}). We feed the image patch embeddings for each frame, $\mathbf{z}_{\text{view}}$ (Eq.~\ref{eq:view_embed}), into a Gaussian splatting decoder to predict point clouds rendered into 2D images.

\paragraph{Video frame and point cloud reconstruction.} For each baseline, we follow the procedure below to decode video frames. For notational simplicity, we denote the video embeddings learned by each baseline as $\mathbf{z} \in \mathbb{R}^{(KT) \times (VSD)}$, where $K$ is the number of trials, $T$ the number of time bins, $V$ the number of views, $S$ the sequence length, and $D$ the embedding dimension. For ResNet AE, $S = 1$, as each view is compressed into a single $D$-dimensional vector. For BEAST and {\sable}, $S = hw$ corresponds to the number of image patches (see Section~\ref{sec:model_architecture}). We set $D = 768$ for all baselines to match the embedding dimension of {\sable}. Since the mouse is head-fixed and the background is static, we subtract the frame average from $\mathbf{z}$ to obtain $\tilde{\mathbf{z}}$, which captures only frame-specific behavioral variations. Since $\tilde{\mathbf{z}}$ is high-dimensional, we apply Principal Component Analysis (PCA) to obtain a compact latent vector $\tilde{\mathbf{z}}_{\text{compact}} \in \mathbb{R}^{(KT) \times P}$ , where $P \ll D$, for decoding. We then predict $\hat{\mathbf{z}}_{\text{compact}}$ from neural activity and reconstruct $\hat{\mathbf{z}} \in \mathbb{R}^{(KT) \times (VSD)}$ via PCA inversion, followed by adding back the frame average. The reconstructed $\hat{\mathbf{z}}$ is then passed through the (pretrained) decoder of the corresponding model, namely ResNet AE, BEAST, or {\sable}, for frame reconstruction~\citep{whiteway2021partitioning}. In addition, {\sable} uniquely enables decoding of 3D animal point clouds across video frames.

\paragraph{Evaluation.} We pretrain all baselines and {\sable} on seven sessions, then apply them zero-shot to extract embeddings from frames corresponding to the 400 sampled trials in each of the three held-out sessions (Section~\ref{sec:data}). These trials are split into 70\% training, 10\% validation, and 20\% test sets.  Following the aforementioned procedure, we compress the extracted video embeddings from each baseline into a $P$-dimensional latent vector ($P = 6$, with 3 principal components (PCs) per view). Given neural activity, we then train a temporal convolutional network (TCN) model \citep{syeda2024facemap} to predict these latents for each baseline (Appendix~\ref{app:tcn_hyperparam}). TCN decoders are trained on the training set, with hyperparameters tuned and checkpoints selected using the validation set. Performance is evaluated on the test set using foreground-only peak signal-to-noise ratio (PSNR) and structural similarity index measure (SSIM) (Appendix~\ref{app:psnr}-\ref{app:ssim}),  computed over the foreground animal segmented using SAM3 \citep{carion2026sam}, where higher values indicate better frame reconstruction quality. See Appendix~\ref{app:foreground_metrics} for details on the rationale and computation of the foreground-only metrics. Hyperparameters are optimized with Ray Tune by sampling 30 configurations (Appendix~\ref{app:tcn_training_details}). This procedure is repeated across all held-out sessions.

\paragraph{Results.} In Fig.~\ref{fig:decoding}, we show that neural activity contains sufficient information to decode high-resolution video frames from the IBL dataset. BEAST outperforms ResNet AE in video frame decoding quality, while {\sable} further surpasses BEAST on both PSNR and SSIM, indicating higher pixel-level fidelity and perceptual similarity to the ground truth (Fig. \ref{fig:decoding}B). All baseline models capture frame-to-frame behavioral variability after accounting for static background information (Fig. \ref{fig:decoding}C). To test whether the neural decoder recovers frame-specific behavior beyond a session-average appearance, we compare its reconstructions with a session-average baseline. This baseline averages {\sable} image patch embeddings across all frames within each session and decodes an image from the averaged embeddings and therefore contains no frame-specific variation. The decoder using {\sable} embeddings significantly outperforms the session-average baseline in PSNR and SSIM (Appendix~\ref{app:decoding_stats}), showing its recovery of frame-to-frame behavioral changes beyond session-average appearance.

More importantly, unlike the 2D baseline methods considered here, {\sable} recovers 3D dynamics, opening new opportunities to study brain–behavior relationships. For example, extending this approach to visual cortex recordings could enable reconstruction of the 3D scene viewed by an animal, helping reveal how neural representations of the environment guide its interactions with the 3D world.


\subsection{Neural Encoding}\label{sec:encoding}
\paragraph{Motivation.} Some behavioral variables, such as wheel speed, can be adequately characterized from 2D video. In contrast, paw movements can include substantial motion toward or away from the camera. Their 2D projections omit depth and depend on viewpoint, so positions and velocities measured in the image may not fully characterize movement in 3D world coordinates. Explicitly modeling 3D behavioral structure may therefore capture information relevant to predicting neural activity that is less accessible in 2D representations. To determine when this additional 3D information is useful, we use neural encoding to evaluate how well the learned behavioral representations predict neural activity. We focus here on the IBL dataset and report Cheese3D results in Appendix~\ref{app:neural_encoding_cheese3d}.

The scientific goal here is to use an encoding model to explain neural activity from behavioral representations and identify \textit{neurons with 3D tuning}: neurons that are more responsive to representations containing 3D geometry, such as depth and shape, than by 2D visual semantic features captured by baselines such as DINOv3 and {\sable} without geometric loss. Both {\sable} and DINOv3 provide two types of video features: [CLS] tokens and image patch embeddings. We use the patch embeddings here as the behavioral representation because {\sable} uses them to predict 3D Gaussian point clouds, encouraging them to encode 3D structure. By comparison, [CLS] tokens summarize global semantic information and are not directly used for 3D reconstruction.


\paragraph{Models.} 
To identify neurons whose activity is sufficiently explained by a 3D behavioral representation, we compare three sets of visual features: (1) image patch embeddings from {\sable}, trained with a 3D geometric loss; (2) patch embeddings from an otherwise matched {\sable} ablation trained without geometric loss, providing a controlled comparison using the same architecture and only 2D reconstruction supervision; and (3) patch embeddings from pretrained DINOv3, providing a reference for 2D semantic features learned through large-scale visual pretraining.

\paragraph{Evaluation.} We benchmark neural encoding by predicting neural activity from behavioral representations extracted by each model. {\sable} and {\sable} trained without a 3D geometric loss are pretrained on seven sessions and then applied zero-shot to extract embeddings from three held-out sessions (Section~\ref{sec:data}); DINOv3 patch embeddings are extracted from the pretrained model. For each held-out session, we train a regularized ridge regression model to predict neural activity from each model's patch embeddings using five-fold cross validation. We  evaluate held-out prediction performance using bits per spike (BPS; Appendix~\ref{app:bps}), with higher values indicating more accurate neural prediction. Results are reported across 2,619 neurons from the three held-out sessions.

\begin{figure}[t]
    \centering
    \includegraphics[width=\textwidth]{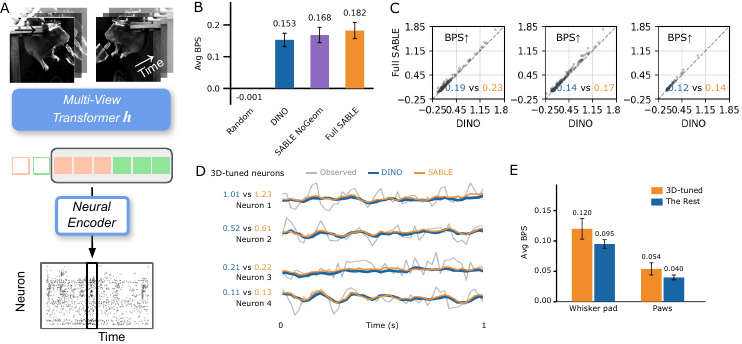}
    \caption{\small {{\bf Neural encoding reveals neurons tuned to 3D behavioral structure beyond 2D  semantics.} 
    (A) Image patch embeddings extracted from each model are used to predict neural activity with a ridge regression neural encoder. 
    (B) Encoding performance is compared across models. DINOv3 provides a strong pretrained visual baseline capturing semantic features from 2D images. {\sable} without geometric loss learns patch embeddings without 3D supervision, whereas full {\sable}'s patch embeddings encode 3D behavioral information. Full {\sable} outperforms all baselines in bits per spike (BPS), while a random-embedding control yields near-zero BPS. Error bars indicate the standard error of the mean. 
    (C) Scatterplot comparison of SABLE vs DINO performance in each held-out session. Each dot corresponds to an individual neuron. The values in the bottom-right corner represent the neuron-averaged BPS per session. Most neurons lie above the diagonal lines, indicating higher BPS when the representations encode 3D behavioral information in addition to 2D semantic features. 
    (D) Comparison of firing rates predicted by SABLE and DINO, averaged over randomly sampled 1-s recording segments, for four example neurons whose firing rates are better predicted by SABLE than by DINO, suggesting greater selectivity for 3D behavioral information. Values above each trace indicate per-neuron BPS.
    (E) Body-part-specific neural encoding analysis of 3D-tuned neurons. The identified 3D-tuned neurons exhibit higher mean BPS than the remaining neurons for both whisker-pad and paw regions, suggesting that 3D behavioral structure in these regions may contribute to explaining their neural activity beyond image-level appearance and behavioral semantics.
     \label{fig:encoding}}}
   
\end{figure}

\paragraph{Results.}
As shown in Fig.~\ref{fig:encoding}B, {\sable} outperforms DINOv3, indicating benefits beyond strong visual pretraining alone. {\sable} also outperforms the matched ablation without geometric loss, supporting the contribution of explicitly learned 3D structure to neural prediction. Fig.~\ref{fig:encoding}C shows that most neurons lie above the diagonal lines, indicating consistent improvements in BPS across neurons and sessions when the representations encode 3D behavioral information in addition to 2D semantic features. At the single-neuron level, we identify neurons whose activity is better predicted by {\sable} than by both DINOv3 and {\sable} without geometric loss across cross-validation folds, which we refer to as exhibiting ``\textit{3D tuning}'' (Fig.~\ref{fig:encoding}D). These neurons encode aspects of behavioral structure that are less fully captured by 2D visual features alone. To further investigate these neurons, we perform body-part-specific neural encoding analysis (Fig.~\ref{fig:encoding}E). {\sable} features extracted from the whisker-pad and paw regions retain 66\% and 28\% of the whole-frame BPS, respectively, indicating that the whisker-pad region alone captures a substantial fraction of the behavioral information relevant to neural prediction. Notably, both regions yield higher mean BPS in 3D-tuned neurons than in the remaining neurons, suggesting that 3D behavioral structure associated with these body parts explain more of the variance in neural activity than 2D behavioral semantics. Identifying these neurons provides candidate populations for investigating how the brain represents 3D posture and movement, and motivates targeted experiments to test whether their neural responses reflect underlying depth information and body geometry rather than image-level appearance and behavioral semantics. These analyses support the interpretability of {\sable} by linking its learned 3D representations to the responses of individual neurons. Beyond improving prediction, they generate testable hypotheses about neurons' sensitivity to 3D behavioral structure, providing a route from behavioral representation learning to neuroscientific insight.

\section{Discussion}
We introduce {\sable}, a self-supervised framework that learns interpretable 3D animal behavior representations from sparse multi-view video. Its geometric inductive biases reconstruct reliable 3D structure where existing methods fail, improving neural encoding and enabling 3D behavior decoding from neural activity. Beyond these methodological advances, we identify neurons with putative 3D tuning whose activity is better explained by {\sable} than by 2D behavioral features, providing candidate populations for investigating how the brain encodes 3D posture and movement. This work has several limitations. First, although {\sable} can be pretrained on multi-animal data and generalize zero-shot to unseen animals, we have only applied it to head-fixed mouse datasets. Extending {\sable} to freely moving animals and other species is an important direction for future work. Second, while {\sable} supports novel-view synthesis, future work could use these synthesized views to extract additional behavioral keypoints and potentially improve neural encoding performance.

\section*{AI use statement}

We used generative AI tools for software debugging, grammar checking, and improving the writing. All AI-assisted content was reviewed by the authors, who take responsibility for the final work.

\section*{Reproducibility statement}

Details on the model hyperparameters, training procedures, and experimental settings are provided in the main text and appendix for reproducibility. The appendix also includes details on the features used, data splits, and preprocessing for both video and neural recordings.

\bibliography{iclr2027_conference}
\bibliographystyle{iclr2027_conference}

\appendix

\section{Datasets}
\subsection{IBL}
\label{app:data_details_ibl}

\subsubsection{\texorpdfstring{PCA + $k$-means}{PCA + k-means} frame selection}
\label{app:frame_selection_ibl}

Since the animal and experimental apparatus remain largely static in IBL videos, uniformly sampling video frames for model training over-represent inactive states and under-represent rare postures. Therefore, we adopt a PCA + $k$-means–based frame-selection scheme to construct a compact, behaviorally diverse training set \citep{wang2026animal}. We adopt the same procedure and select approximately 1,000 frames per session for model training. For each session, frames from the left-camera video are resized to a low spatial resolution, vectorized, and used to compute frame-wise motion energy via absolute pixel-wise differences between consecutive frames. We retain frames above a motion-energy percentile threshold, fit PCA within each session on the resulting high-motion frame vectors, and apply $k$-means clustering to identify anchor frames. Each anchor index $i$ is then expanded to $\{i-1, i, i+1\}$ to incorporate short temporal context. The resulting frame indices are reused for the right camera to ensure that the extracted video frames are temporally synchronized across the left and right views. In total, 1,000 synchronized frame pairs are selected from each session for training. For model evaluation (neural encoding and decoding), we extract video frames temporally aligned with the neural data (Section~\ref{sec:data}). Each session contains 400 randomly sampled 1-second time segments, with videos sampled at 60 Hz, yielding 24,000 aligned frames from each view. This separation allows the baseline models to be trained on a compact yet diverse set of unlabeled video frames, while neural analyses are performed on frames corresponding to the sampled neural time segments.

\subsubsection{Lightning Pose keypoints}\label{app:lp_pose_ibl}

Keypoints are obtained using the Lightning Pose 3D (LP3D) framework \citep{aharon2026lightning}. We train an ensemble of three LP3D models on 200 manually labeled frames per session using distinct random seeds for the train/validation split. Each model uses a ViT-S/DINO vision transformer backbone within a multi-view transformer architecture, with patch masking and a 3D reprojection loss that leverages camera calibration parameters to enforce geometric consistency across views during training. We then run inference on all video frames for each ensemble member.

To obtain post-processed predictions with calibrated uncertainty estimates, we apply the nonlinear multi-view Ensemble Kalman Smoother (mvEKS) with variance inflation to the ensemble outputs. This post-processing step jointly exploits temporal continuity and multi-view geometric constraints, inflating observation noise for views whose predictions are inconsistent with the remaining views and thereby downweighting overconfident erroneous predictions.

To produce final keypoint trajectories suitable for large-scale analysis, we apply a distillation procedure \citep{aharon2026lightning}. Frames are first filtered by retaining the $\sim$60\% with the lowest posterior predictive variance from the nonlinear mvEKS, keeping only high-confidence predictions. The retained frames are then filtered for pose diversity using $k$-means clustering in 3D pose space ($k=3000$). This procedure generates high-quality and diverse pseudo-labels. The resulting pseudo-labeled frames are combined with the original labeled frames across all seeds to form an augmented training set. A single LP3D model is then retrained on this combined dataset and applied to the full videos to generate the final keypoint trajectories used in all downstream analyses.

\subsubsection{Video depth extraction}
\label{app:vda_depth_ibl}
\sable{} uses monocular depth estimates as an auxiliary geometric signal for sparse-view 3D reconstruction (Section~\ref{sec:model_architecture}). Monocular relative depth is extracted using Video Depth Anything (VDA) \citep{chen2025videodepthanything}, which is used as a frozen data preprocessing model rather than a trainable model component. We choose VDA for depth estimation, because it produces temporally consistent depth predictions across video frames. We use the publicly released ViT-B checkpoint and apply it independently to the extracted left- and right-camera RGB videos for each session to generate dense single-channel relative depth maps. We use relative rather than metric depth, since absolute scene scale is not required for our reconstruction objective and relative depth provides sufficient geometric structure for regularization. In the main experiments, RGB images are resized to $320\times320$, and the corresponding depth maps are resized to the same spatial resolution to preserve pixel alignment between RGB and depth representations.

\subsection{Cheese3D: Novel-View Synthesis and Neural Encoding}
\label{app:data_details_cheese3d}

\begin{figure}[ht]
    \centering
    \includegraphics[width=\textwidth]{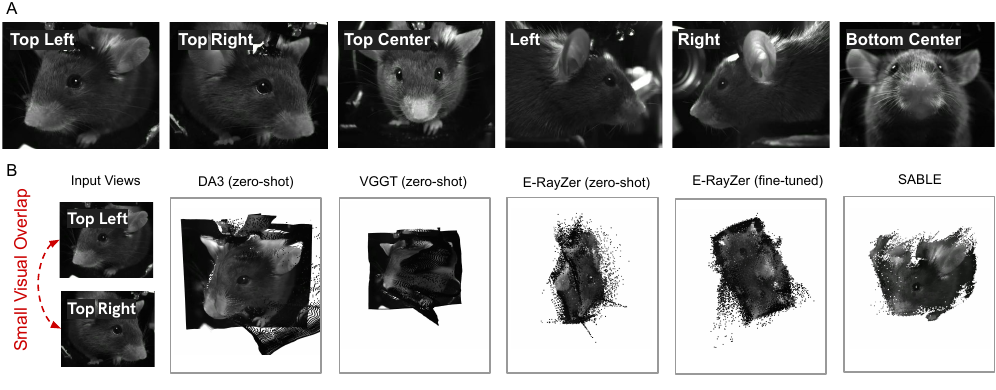}
    \caption{\small{\bf Cheese3D video frames and 3D reconstruction comparison.} (A) Six camera views from the Cheese3D dataset. (B) 3D reconstruction from the selected sparse view pair (top-right and top-left) of Cheese3D. {\sable} reliably reconstructs 3D animal behavior, while existing 3D foundation models exhibit distinct failure modes, consistent with the results on IBL (Fig.~\ref{fig:3d_recon_ibl}). DA3 \citep{lin2025depth} and VGGT \citep{wang2025vggt} produce plausible per-view point clouds; however, DA3 fails to consistently align the two point clouds, with noticeable misregistration around the nose, and VGGT fails to disentangle foreground and background in depth, leaving background geometry intermingled with the mouse. Both zero-shot and fine-tuned E-RayZer \citep{zhao2026erayzer} collapse to degenerate, flattened 3D structures. In contrast, {\sable} recovers consistent cross-view geometry, separates foreground from background in depth, and avoids degenerate solutions, producing a coherent 3D reconstruction of the mouse.}
    
\label{fig:cheese3d_view_demos}
\end{figure}

\subsubsection{Dataset}
\label{app:cheese3d_data}
The Cheese3D dataset consists of six synchronized camera views of head-fixed mice, capturing whole-face movements at 100 Hz with a spatial resolution of $1280\times1024$ pixels (see Fig.~\ref{fig:cheese3d_view_demos} panel A). For our experiments, we downsample the videos to 60 Hz. The dataset contains 11 video-recording sessions, one of which includes simultaneously recorded neural activity. We use the 10 sessions without neural recordings for model training, using approximately 5,500 synchronized frame pairs, and use the session with neural recordings for neural encoding evaluation.

To evaluate whether {\sable}'s sparse-view reconstruction capability generalizes beyond the IBL dataset, we construct an analogous two-view setting from Cheese3D. Specifically, we select the top-left and top-right camera views, which have limited overlap and therefore provide a challenging sparse-view configuration similar to that used for IBL. {\sable} reconstructs coherent 3D mouse structure from the selected sparse view pair. In contrast, existing 3D foundation models produce inconsistent or degenerate reconstructions, exhibiting distinct failure modes (Fig.~\ref{fig:cheese3d_view_demos}, panel B).

\subsubsection{Lightning Pose keypoints}
\label{app:lp_pose_cheese3d}

Similar to IBL frame extraction, to select pose-diverse video frames from Cheese3D, we adopt a keypoints + $k$-means strategy \citep{wang2026beast3danimalbehavioralanalysis}. We first obtain geometrically consistent keypoints across all six camera views using a frame-filtering and triangulation procedure. To be more specific, we train an ensemble of three single-view Lightning Pose 3D (LP3D) models \citep{aharon2026lightning} on 665 manually labeled instances. Each instance corresponds to a time point and contains labels from a subset of the camera views in which the facial keypoints are anatomically visible, out of up to six camera views. We then run the ensemble on the full-length videos from all views and retain only frames for which the median keypoint likelihood exceeds 0.6 in at least two views. For each retained frame, we triangulate the keypoints across the available views using the calibrated camera parameters to obtain their 3D locations. Finally, we project the triangulated 3D keypoints back into all six camera views, yielding geometrically consistent 2D keypoint estimates for subsequent pose-diverse frame selection and quality control.

\subsubsection{Keypoints + \texorpdfstring{$k$-means}{k-means}}
\label{app:frame_selection_cheese3d}
We next select a pose-diverse subset of frames. For each session, the triangulated 3D keypoints are flattened into a single pose vector, and $k$-means clustering is applied in this pose space. For each cluster, we select the frame whose 3D pose is closest to the cluster centroid, yielding approximately $50-55$ representative frames per session. 

We then perform an additional geometric quality-control step based on reprojection error. Specifically, for each selected frame, we compare the median LP3D 2D keypoint predictions with the 2D locations obtained by projecting the triangulated 3D keypoints back into each camera view. The reprojection error is defined as the Euclidean distance between these two estimates. We discard the 25\% of selected frames with the largest reprojection errors, thereby removing frames whose multi-view predictions are least geometrically consistent. The remaining frames are used to construct the final training and test sets, containing 450 and 150 instances, respectively.

\subsubsection{Video depth extraction}
\label{app:vda_cheese3d}
To obtain monocular depth estimates used by \sable{}, we apply Video Depth Anything (VDA) using the same preprocessing procedure described for IBL in Section~\ref{app:vda_depth_ibl}. Specifically, VDA is applied independently to the Cheese3D RGB videos, with input images non-uniformly resized to $320\times320$ pixels without cropping, producing temporally consistent relative-depth maps aligned with the corresponding RGB frames. 

\begin{figure}[h!]
    \centering
    \includegraphics[width=0.75\textwidth]{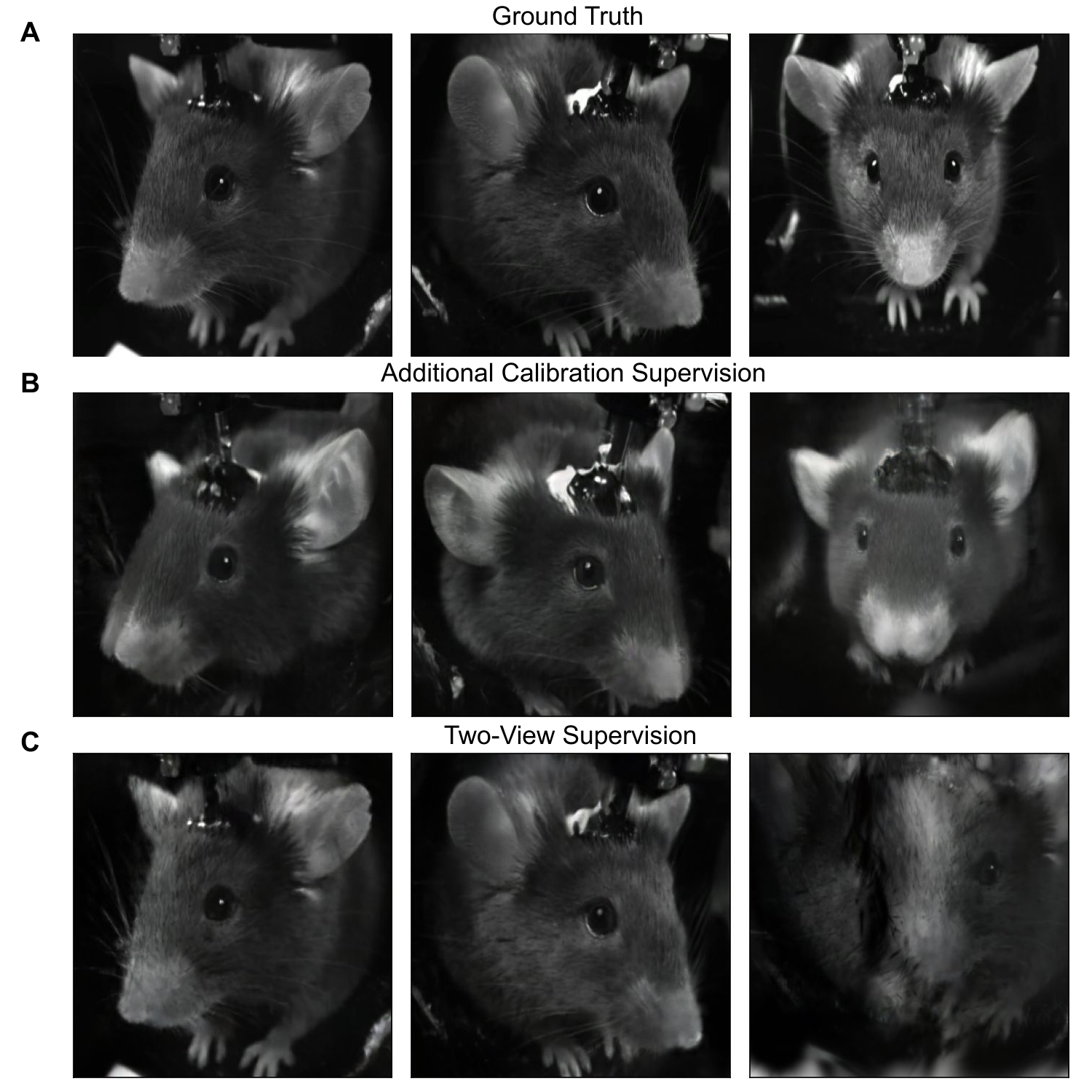}
        \caption{\small{\bf Novel view reconstruction on Cheese3D.} (A) Ground-truth top-left and top-right views (model inputs) and the top-center view (held out). (B) 2D reconstructions from {\sable} using fixed camera calibration during training. The reconstructed held-out view is clear. (C) 2D reconstructions from {\sable} using inferred camera poses during training. The mouse's face is visible in the reconstructed held-out view but is blurrier than in panel (B). Both panels (B) and (C) show that {\sable} recovers meaningful 3D structure of the animal.}    
\label{fig:cheese3d_nvs}
\end{figure}

\begingroup

\subsubsection{Novel View Reconstruction}\label{app:nvs_cheese3d}
Quantitative evaluation of 3D reconstruction on IBL data is challenging because it provides only two camera views, leaving no additional viewpoint for validation. We therefore evaluate {\sable} on Cheese3D, which provides additional views and corresponding camera parameters. Because Cheese3D lacks ground-truth 3D labels, we assess reconstruction indirectly through novel-view synthesis (NVS), comparing predicted views with observed views from a held-out camera. This evaluation reflects both reconstruction quality and camera calibration accuracy.

\paragraph{Evaluation.}
We compare \sable{} with E-RayZer \citep{zhao2026erayzer}. Each model receives only the top-left and top-right views and predicts a 3D point cloud together with camera poses. We render the held-out top-center view using its inferred camera pose and compare the rendering with the corresponding GT image using PSNR and SSIM (mean $\pm$ 1 SE across 100 test frames). Although the inferred camera pose is close to the ground truth camera poses, errors in the estimated input camera poses can affect the reconstructed geometry and its alignment with the evaluation camera.

\paragraph{Results.}
Table~\ref{tab:nvs_results} reports scores for the held-out top-center view and the two input views, with an all-black image as a reference control. \sable{} outperforms E-RayZer on both held-out-view metrics, showing stronger novel-view reconstruction from only two input views. Although SABLE recovers meaningful 3D animal shapes (Appendix Figure~\ref{fig:cheese3d_nvs} B), novel-view PSNR and SSIM remain modest in Table \ref{tab:nvs_results}, which suggest remaining reconstruction or camera-calibration
errors, which we investigate below.

\begin{table}[!ht]
\centering
\small
\caption{Novel-view synthesis results on Cheese3D using two-view supervision, comparing E-RayZer and \sable{} on input views and a held-out top-center view. Values are mean $\pm$ 1 SE across 100 test frames.}
\vspace{+1mm}
\label{tab:nvs_results}
\begin{tabular}{lccc}
\hline
\multicolumn{4}{c}{PSNR $\uparrow$} \\ \hline
View & All-Black Image & E-RayZer & \sable{} \\ \hline
Top-Left & 11.31 & $23.69 \pm 0.33$ & $\mathbf{29.74 \pm 0.06}$ \\
Top-Right & 11.88 & $28.10 \pm 0.36$ & $\mathbf{28.78 \pm 0.09}$ \\
Top-Center (Held-Out) & 7.35 & $10.74 \pm 0.10$ & $\mathbf{12.74 \pm 0.06}$ \\
\hline
\multicolumn{4}{c}{SSIM $\uparrow$} \\ \hline
View & All-Black Image & E-RayZer & \sable{} \\ \hline
Top-Left & 0.004 & $0.812 \pm 0.009$ & $\mathbf{0.892 \pm 0.001}$ \\
Top-Right & 0.004 & $0.863 \pm 0.008$ & $\mathbf{0.898 \pm 0.002}$ \\
Top-Center (Held-Out) & 0.001 & $0.310 \pm 0.004$ & $\mathbf{0.362 \pm 0.002}$ \\
\hline
\end{tabular}
\end{table}

\paragraph{Effect of camera calibration.}
To investigate the contribution of camera estimation, we compare the original two-view model with a variant that receives additional top-center camera pose calibration supervision during training. Both variants receive only top-left and top-right image tokens. For the new variant, the top-center view is fully masked, and we only provide the fixed ground-truth camera parameters during training. We evaluate both variants on 520 held-out top-center frames, with camera parameters fixed. As shown in Table~\ref{tab:nvs_calibration}, additional calibration supervision improves PSNR by 6.43 dB and SSIM by 0.281 while using the same two input views. These results suggest that the scores in Table \ref{tab:nvs_results} partly reflect the difficulty of estimating accurate camera poses from two views, even when meaningful 3D shapes are recovered. 

\begin{table}[!ht]
\centering
\small
\caption{Effect of additional camera calibration supervision on \sable{} reconstruction quality. Both variants receive the image tokens from the same two input views and are evaluated on 520 held-out top-center frames. Values are mean $\pm$ 1 SE across frames.}
\vspace{+1mm}
\label{tab:nvs_calibration}
\begin{tabular}{lcc}
\hline
Training Setting & PSNR $\uparrow$ & SSIM $\uparrow$ \\
\hline
Two-View Supervision
& $12.093 \pm 0.031$
& $0.3821 \pm 0.0010$ \\
Additional Calibration Supervision
& $\mathbf{18.519 \pm 0.032}$
& $\mathbf{0.6633 \pm 0.0008}$ \\
\hline
\end{tabular}
\end{table}

\par
\endgroup

\subsubsection{Neural Encoding}
\label{app:neural_encoding_cheese3d}

To evaluate whether {\sable}'s behavioral representations support neural prediction beyond the IBL dataset, we evaluate {\sable} and the baseline models on Cheese3D under a sparse-view setting. Each  model is trained using two camera views. For neural encoding, we concatenate a frame-level embedding from each view and use them to predict neural activity with a TCN encoder (Appendix~\ref{app:tcn_hyperparam}). These embeddings are [CLS] tokens for BEAST, DINOv3 [CLS] tokens for {\sable}, and latent features for ResNet AE. This benchmark evaluates neural prediction quality using global, semantic visual features.

\paragraph{View selection.}
\label{app:view_selection_cheese3d}
We select the top-left and top-right Cheese3D views, which have limited visual overlap, analogous to the two-view configuration used for IBL.

\paragraph{Foreground mask extraction.}
\label{app:sam3_cheese3d}
When merging the per-view point clouds directly, background and mouse surfaces can intersect. To isolate the animal surface from the background, we generate a foreground mouse mask using SAM3 \citep{carion2026sam} following \citep{wang2026beast3danimalbehavioralanalysis}. We use this mask in two ways. First, we assign background pixels the farthest normalized depth before constructing the two-view point cloud, thereby preventing background pixels from producing spurious points near the mouse surface. Second, we compute $\mathcal{L}_{\text{recon}}$ and $\mathcal{L}_{\text{geom}}$ only over foreground pixels. This prevents the background from contributing to these objectives and focuses the optimization on reconstructing the mouse. 

\paragraph{Rigid Transformation.}
\label{app:affine_transformation_cheese3d}
For Cheese3D, we manually determine a single rigid transformation that is applied consistently across sessions to align the two-view point clouds and construct the pseudo point cloud used for loss regularization. No LightningPose keypoints are used for this alignment. Despite this simplified alignment procedure, \sable{} recovers the underlying 3D structure (Fig.~\ref{fig:cheese3d_view_demos}).

\paragraph{Neural Encoding Model Traiåning.}
\label{app:encoding_model_cheese3d}
For neural encoding, models are trained on 10 sessions using approximately 5,500 synchronized frame pairs and evaluated on a held-out session. The held-out session contains simultaneously recorded electrophysiological activity measured with a 32-channel single-shank silicon probe in the brainstem. We randomly sample 1,233 one-second trials from the recording session. Neural activity is available from 8 neurons and is subsequently binned at 60 Hz, matching the downsampled video frame rate, resulting in one neural activity vector per video frame.

\begin{table}[ht!]
\centering
\small
\caption{Neural encoding results on the Cheese3D dataset (TCN encoder; mean $\pm$ 1 standard error across 8 neurons).}
\label{tab:encoding_cheese3d}
\begin{tabular}{l*{3}{c}}
\hline
 & ResNet AE & BEAST & {\sable} \\ \hline
Encoding (bps) & 0.603 $\pm$ 0.144 & 0.452 $\pm$ 0.088 & \textbf{0.655 $\pm$ 0.146} \\
\hline
\end{tabular}
\end{table}

\paragraph{Neural Encoding Results.}
\label{app:encoding_results_cheese3d}
Table~\ref{tab:encoding_cheese3d} reports mean BPS $\pm$ one standard error across the 8 recorded neurons in the held-out session. {\sable} achieves the highest mean BPS, providing additional evidence of improved neural prediction beyond the IBL dataset.

\section{Metrics and Baselines}
\label{app:metrics}

\subsection{Bits per spike}\label{app:bps}
Bits per spike (BPS) quantifies the improvement in log-likelihood of a predictive model over a null (mean-rate) model, normalized by the number of observed spikes and expressed in base-2 units. For neuron $n$, it is defined as
\begin{equation}
    \mathrm{BPS}
    = \frac{\mathcal{L}_{\mathrm{null}} - \mathcal{L}_{\mathrm{pred}}}
           {N_{\mathrm{sp}}\,\ln 2}.
\end{equation}
A higher BPS indicates that the model assigns higher likelihood to the observed spikes than the null model, with $\mathrm{BPS}=0$ corresponding to the mean-rate baseline.

\subsection{Random-embedding baseline for neural encoding}
\label{app:random_embedding_baseline}
A BPS value of 0 corresponds to a model that collapses to predicting the trial-averaged neural activity and fails to capture variability across individual trials. We expect a random model to perform at or near this trial-averaged baseline because it does not contain information about the animal behavior. The models’ positive BPS values suggest that they capture behavior-related information from videos that is predictive of neural activity, rather than simply reproducing the average neural response. 

As a null baseline, we pool embeddings from DINOv3 [CLS] tokens of {\sable} across all trials and time bins and randomly permute them, thereby destroying the temporal structure while preserving the embedding content within each frame. As shown in figure \ref{fig:encoding}, the random baseline only achieves near chance-level (0 bps) encoding performance, and therefore, proves {\sable} and beseline models capture variability across trials.

\subsection{Peak signal-to-noise ratio (PSNR)}\label{app:psnr}

Peak signal-to-noise ratio (PSNR) measures pixel-level reconstruction fidelity. For a ground-truth image $x\in[0,M]^{H\times W\times C}$ and a reconstruction $\hat{x}$, we first compute the mean squared error
\begin{equation}
    \mathrm{MSE}(x,\hat{x})
    =
    \frac{1}{HWC}
    \sum_{h=1}^{H}
    \sum_{w=1}^{W}
    \sum_{c=1}^{C}
    \left(x_{hwc}-\hat{x}_{hwc}\right)^2.
\end{equation}
The PSNR is then
\begin{equation}
    \mathrm{PSNR}(x,\hat{x})
    =
    10\log_{10}
    \left(
        \frac{M^2}{\mathrm{MSE}(x,\hat{x})}
    \right),
\end{equation}
where $M$ denotes the maximum possible pixel value. In our experiments, we normalize images to $[0,1]$, yielding $M = 1$. Higher PSNR corresponds to lower reconstruction error in the mean-squared sense. Since PSNR is a monotonic transformation of MSE, it primarily reflects pixel-wise accuracy and is most appropriate for assessing low-level reconstruction quality.

\subsection{Structural similarity index measure (SSIM)}\label{app:ssim}

Structural similarity index measure (SSIM) evaluates reconstruction quality by comparing local luminance, contrast, and structural statistics \citep{wang2004image}. Given a local window extracted from a ground-truth image $x$ and a reconstruction $\hat{x}$, SSIM is defined as
\begin{equation}
\mathrm{SSIM}(x,\hat{x})
=
\frac{
(2\mu_x\mu_{\hat{x}} + C_1)
(2\sigma_{x\hat{x}} + C_2)
}{
(\mu_x^2 + \mu_{\hat{x}}^2 + C_1)
(\sigma_x^2 + \sigma_{\hat{x}}^2 + C_2)
},
\end{equation}
where $\mu_x$ and $\mu_{\hat{x}}$ denote local means, $\sigma_x^2$ and $\sigma_{\hat{x}}^2$ are local variances, $\sigma_{x\hat{x}}$ is the local covariance, and $C_1, C_2$ are stabilization constants.

We compute SSIM using $11\times11$ Gaussian windows and average over spatial locations and channels. Compared to PSNR, SSIM is designed to better align with perceptual image quality, assigning higher scores to reconstructions that preserve local structure, edges, and contrast, even when pixel-wise errors are similar.

\subsection{Foreground-only PSNR and SSIM}\label{app:foreground_metrics}

In the neural decoding experiment, we provide a direct evaluation of behavior-related visual reconstruction quality. Nevertheless, among baseline models’ decoders may affect reconstruction quality in static background regions, which could confound PSNR and SSIM. Therefore, we compare the reconstruction image quality by computing both metrics exclusively on the mouse foreground.

Specifically, we use SAM3 \citep{carion2026sam} to generate mouse foreground masks of the ground truth frames using the text prompt ‘mouse’, and then we apply the foreground masks on ground truth frames and the reconstructed frames, yielding frames with uniformly black background for all baseline reconstructions. We then compute PSNR and SSIM on the masked frames. 

\begingroup

\subsection{Session-average baseline for decoding}
\label{app:decoding_stats}

\paragraph{Session-average baseline and statistical tests.}
To test whether neural decoding recovers frame-specific behavior beyond average appearance, we decode the session-average embedding at every test time point, without neural input. The baseline and neural-decoded reconstructions use the same {\sable} decoder, test frames, foreground masks, and evaluation metrics (Appendix~\ref{app:foreground_metrics}), differing only in their input embeddings. Table~\ref{tab:decoding_mean_baseline_abs} reports performance for each held-out session. We average PSNR and SSIM within each trial across time bins and camera views, then compute paired differences ({\sable} minus the session-average baseline). We use one-sided paired sign-flip permutation tests with $10^5$ permutations to test whether the mean difference is greater than zero. We obtain two-sided 95\% percentile confidence intervals from $10^5$ bootstrap resamples of trials within each session. 

\paragraph{Results.}
Fine-grained paw and whisker movements occupy small image regions, so improvements in reconstructing them may contribute only modestly to image-averaged metrics such as PSNR and SSIM. Nevertheless, neural decoding using {\sable} embeddings outperforms the session-average baseline across all three sessions (one-sided paired sign-flip tests, $p<10^{-4}$), supporting recovery of frame-specific behavioral information beyond session-average appearance.
\par
\endgroup

\begin{table}[ht!]
\centering
\small
\caption{Mean foreground PSNR and SSIM of {\sable} reconstructions decoded from neural activity and from the session-average baseline, computed on test trials from each held-out session. For each session, the average metrics across all test frames are shown.}
\label{tab:decoding_mean_baseline_abs}
\begin{tabular}{lcccc}
\hline
 & \multicolumn{2}{c}{PSNR $\uparrow$}
 & \multicolumn{2}{c}{SSIM $\uparrow$} \\
Session & Session-average & {\sable} & Session-average & {\sable} \\
\hline
4b00df29 & 23.73 & 24.18 & 0.924 & 0.925 \\
72cb5550 & 21.49 & 22.55 & 0.898 & 0.903 \\
781b35fd & 19.40 & 20.33 & 0.845 & 0.850 \\
\hline
Mean & 21.54 & 22.35 & 0.889 & 0.893 \\
\hline
\end{tabular}
\end{table}

\section{Ablation Study}

\subsection{Model Size}\label{app:model_size_ablation}

We benchmark {\sable} against a lightweight CNN autoencoder (ResNet-AE-18) and a ViT masked autoencoder (BEAST) for neural decoding (Section~\ref{sec:decoding}) using a TCN encoder. \citet{wang2026animal} showed that BEAST outperforms the larger ResNet-AE-50 on downstream pose estimation, motivating our choice of BEAST as the primary high-capacity baseline.

However, model size may independently affect decoding performance. We therefore perform a model-size ablation on the IBL dataset, evaluating larger variants of both ResNet-AE (ResNet-152) and BEAST (ViT-Large) using the same neural decoding benchmark. As shown in Table~\ref{tab:model-size-ablation}, increasing model size does not improve decoding performance for either baseline. {\sable} achieves the highest decoding PSNR and SSIM across all models. These results indicate that {\sable}'s decoding advantage is not simply explained by scaling the backbone, highlighting the importance of the representation learning architecture in this setting.

\begin{table}[ht!]
\centering
\small
\caption{Neural decoding performance (PSNR, SSIM) across model sizes on the IBL dataset.}
\label{tab:model-size-ablation}
\begin{tabular}{lcc}
\hline
Method & Decoding (PSNR) $\uparrow$ & Decoding (SSIM) $\uparrow$ \\ \hline
ResNet-AE-18 & 21.013 $\pm$ 0.015 & 0.853 $\pm$ 0.000 \\
ResNet-AE-152 & 19.520 $\pm$ 0.017 & 0.840 $\pm$ 0.000 \\
BEAST (ViT Base) & 21.659 $\pm$ 0.009 & 0.858 $\pm$ 0.000 \\
BEAST (ViT Large) & 21.210 $\pm$ 0.008 & 0.852 $\pm$ 0.000 \\
{\sable} & \textbf{22.355 $\pm$ 0.014} & \textbf{0.893 $\pm$ 0.000} \\
\hline
\end{tabular}
\end{table}

\subsection{Loss Weighting}\label{app:loss_weighting_ablation}

The loss weights were selected through preliminary hyperparameter tuning and consistently produced high-quality 2D renderings and 3D reconstructions. For fair comparison, we similarly tuned the ResNet-AE and BEAST baselines to achieve strong 2D reconstruction performance before downstream evaluation.

We ablated the three loss weights by varying $\mathcal{L}_{\text{recon}}$, $\mathcal{L}_{\text{percep}}$, and $\mathcal{L}_{\text{geom}}$ with all other settings fixed. In Table~\ref{tab:loss_weighting_ablation}, we report the PSNR of the reconstructed view. The model performance is generally robust to moderate changes in the loss weights. Although removing $\mathcal{L}_{\text{geom}}$ or increasing $\mathcal{L}_{\text{recon}}$ slightly improves PSNR, the former causes the reconstructed 3D point clouds to collapse, while the latter introduces noticeable geometric distortions by overemphasizing 2D reconstruction. Among the loss combinations that preserve coherent 3D geometry, our chosen weighting achieves the highest PSNR.

\begin{table}[ht!]
\centering
\small
\caption{Ablation on loss weighting.}
\label{tab:loss_weighting_ablation}
\begin{tabular}{lcccc}
\hline
Variant & $\mathcal{L}_{\text{recon}}$ & $\mathcal{L}_{\text{percep}}$ & $\mathcal{L}_{\text{geom}}$ & PSNR $\uparrow$ \\ \hline
\textbf{Ours}    &\textbf{1.0} &\textbf{0.3} &\textbf{1.0} &\textbf{26.0} \\
No $\mathcal{L}_{\text{geom}}$    &1.0 &0.3 &0.0 &26.9 \\
High $\mathcal{L}_{\text{recon}}$ &2.0 &0.3 &1.0 &26.1 \\
Low $\mathcal{L}_{\text{percep}}$ &1.0 &0.1 &1.0 &25.9 \\
No $\mathcal{L}_{\text{percep}}$  &1.0 &0.0 &1.0 &25.7 \\
High $\mathcal{L}_{\text{percep}}$&1.0 &1.0 &1.0 &24.7 \\
High $\mathcal{L}_{\text{geom}}$  &1.0 &0.3 &2.0 &25.6 \\
Low $\mathcal{L}_{\text{recon}}$  &0.5 &0.3 &1.0 &25.2 \\
\hline
\end{tabular}
\end{table}

\section{Model and hyperparameter details}
\label{app:model_details}

\subsection{\sable} \label{app:model_hyperparams}

{\sable} builds upon the E-RayZer backbone~\citep{zhao2026erayzer}, and we use the \href{https://huggingface.co/qitaoz/E-RayZer/blob/main/checkpoints/erayzer_multi.pt}{\texttt{erayzer\_multi.pt}} pretrained checkpoint available on Hugging Face. Unlike the original E-RayZer model, which adopts cross-view masking, {\sable} applies patch-level masking across both views. We set the loss weights in Eq.~\ref{eq:model_formula} to $\lambda_{\text{recon}} = 1$, $\lambda_{\text{percep}} = 0.3$ and $\lambda_{\text{geom}} = 1$. Table~\ref{tab:sable_model_hparams} summarizes the model hyperparameters used in the main experiments.

\begin{table}[htbp]
\centering
\small
\caption{\sable{} model hyperparameters.}
\label{tab:sable_model_hparams}
\begin{tabular}{p{0.38\linewidth}p{0.54\linewidth}}
\hline
Hyperparameter & Value \\ \hline
Input image size & $320 \times 320$ \\
Patch Size & $16$ \\
Number of Image Tokens & $400$ \\
Embedding Dimension & $768$ \\
DINOv3 encoder & ViT-B/16 \\
Number of Attention Heads & $12$ \\
Image Encoder Depth & $16$ \\
Geometry Encoder Depth & $16$ \\
Decoder Depth & $8$ \\
Neck Depth & $1$ \\
Mask Ratio & $0.5$ for IBL; 0.75 for Cheese3D \\
Gaussian Spherical Harmonics Degree & $3$ \\
Near Plane & $0.2$ \\
Depth Range & $(0, 8)$ \\
Scene Scale Factor & $1.35$ \\
\hline
\end{tabular}
\end{table}

\subsection{ResNet autoencoder}
\label{app:resnet_ae_hyperparams}
The ResNet autoencoder baseline \citep{he2016deep} compresses each video frame into a low-dimensional latent vector and reconstructs the input image from this latent representation. The model uses a ResNet-based convolutional encoder followed by an MLP bottleneck and a symmetric decoder for image reconstruction. Table~\ref{tab:resnet_ae_model_hparams} summarizes the model hyperparameters used in the main experiments.

\begin{table}[htbp]
\centering
\small
\caption{ResNet autoencoder model hyperparameters.}
\label{tab:resnet_ae_model_hparams}
\begin{tabular}{p{0.38\linewidth}p{0.54\linewidth}}
\hline
Hyperparameter & Value \\ \hline
Input image size & $224\times224$ \\
Image preprocessing & ImageNet normalization \\
Backbone & ResNet-18 \\
Residual block configuration & $(2,2,2,2)$ \\
Embedding Dimension & 768 \\
Bottleneck mapping & $512\times7\times7 \rightarrow 768 \rightarrow 512\times7\times7$ \\
Batch Size & 32 \\
\hline
\end{tabular}
\end{table}

\subsection{BEAST}
\label{app:beast_hyperparams}
The BEAST baseline is built on a ViT-MAE architecture with a ViT-B backbone pretrained on ImageNet and learns video embeddings through masked image reconstruction and temporal contrastive learning. The model tokenizes each frame into image patches with a patch size of 16, applies random patch masking with a standard mask ratio of 75\%, and reconstructs the masked regions using a mean squared error (MSE) reconstruction loss. In addition, a frame-level contrastive (InfoNCE) loss encourages temporal consistency between nearby video frames. Table~\ref{tab:beast_model_hparams} summarizes the model architecture and hyperparameters used in our experiments.

\begin{table}[ht!]
\centering
\small
\caption{BEAST model hyperparameters.}
\label{tab:beast_model_hparams}
\begin{tabular}{p{0.38\linewidth}p{0.54\linewidth}}
\hline
Hyperparameter & Value \\ \hline
Input image size & $224\times224$ \\
Patch size & $16$ \\
Embedding Dimension & $768$ \\
Encoder layers & $12$ \\
Attention heads & $12$ \\
MLP intermediate size & $3072$ \\
Activation & GELU \\
QKV bias & Enabled \\
Decoder hidden dimension & $512$ \\
Decoder layers & $8$ \\
Decoder attention heads & $16$ \\
Decoder intermediate size & $2048$ \\
MAE mask ratio & $0.75$ \\
Projection / embedding size & $768$ \\
Batch Size & 32 \\
\hline
\end{tabular}
\end{table}


\subsection{TCN}
\label{app:tcn_hyperparam}
For neural decoding and the [CLS]-token-based neural encoding analyses in Appendices~\ref{app:neural_encoding_cheese3d} and~\ref{app:model_size_ablation}, we use the temporal convolutional network (TCN) architecture from the Facemap study~\citep{syeda2024facemap}. The model processes inputs (behavioral features for encoding or neural activity for decoding) through a hierarchy of temporal convolutional layers with multiple receptive-field scales to capture temporal dynamics across different timescales. The resulting temporal features are passed through feedforward layers to produce compact latent representations, followed by a linear readout layer that predicts neural activity for encoding or behavioral features for decoding. Table~\ref{tab:tcn_model_hparams} summarizes the model architecture and hyperparameters used in our experiments.

\begin{table}[ht!]
\centering
\small
\caption{TCN model hyperparameters.}
\label{tab:tcn_model_hparams}
\begin{tabular}{p{0.38\linewidth}p{0.54\linewidth}}
\hline
Hyperparameter & Value \\ \hline
Input dimension & Neuron count (decoding) or behavioral feature dimension (encoding) \\
Temporal filters & $10$ \\
Temporal kernel size & $201$ \\
Core layers & $2$ \\
Intermediate dimension & $50$ \\
Latent dimension & $256$ \\
Readout layers & $1$ \\
Output dimension & Neuron count (encoding) or behavioral feature dimension (decoding) \\
Temporal filter initialization & Gabor Wavelets \\
Wavelet nonlinearity & ReLU \\
Latent nonlinearity & ReLU \\
\hline
\end{tabular}
\end{table}


\section{Training details}
\label{app:training_details}

\subsection{\sable}
\label{app:sable_training}
{\sable} is initialized from a pretrained E-RayZer checkpoint and further pretrained on the IBL two-camera dataset (approximately 7,000 frame pairs from 7 sessions) and the Cheese3D dataset (5,500 frame pairs from 10 sessions), respectively. Unless otherwise stated, each training sample consists of two synchronized camera views and corresponding reconstruction targets. The model jointly infers camera geometry and reconstructs 3D structure through a Gaussian splatting decoder conditioned on multi-view transformer embeddings. Training data from each session are split into training and validation subsets using a 90\% and 10\% split. The \sable{} training objective combines image reconstruction, perceptual, and geometric consistency losses. Optimization is performed using AdamW with a learning rate of $2\times10^{-5}$, weight decay of $0.05$, batch size 24 for IBL and 16 for Cheese3D, and a cosine learning-rate schedule with 15\% warmup steps and a cosine division factor of 10. Training is performed for 100 epochs on a single NVIDIA A40 or A100 GPU and takes approximately 17 hours per session.

\subsection{ResNet autoencoder}
\label{app:resnet_ae_training}
The ResNet autoencoder is trained on the same video frame pairs from the IBL and Cheese3D datasets, as described in Appendix~\ref{app:frame_selection_ibl} and Appendix~\ref{app:neural_encoding_cheese3d}, respectively. We use an Adam optimizer with a learning rate of $2\times10^{-5}$, batch size 24, weight decay of $0.05$, and a cosine learning-rate schedule with 15\% warmup steps and a cosine division factor of 10. Training uses the default image augmentation pipeline from the \href{https://github.com/paninski-lab/beast}{BEAST
 code repository} and runs for 160 epochs for IBL and 400 for Cheese3D. Training is performed on a single NVIDIA A100 GPU and takes approximately 0.5 hours per session.

\subsection{BEAST}
\label{app:beast_training}
The BEAST backbone baseline is trained on the same video frame pairs from the IBL and Cheese3D datasets, as described in Appendix~\ref{app:frame_selection_ibl} and Appendix~\ref{app:neural_encoding_cheese3d}, respectively. We use a standard mask ratio of 0.75 for training the ViT-MAE backbone. Optimization is performed using Adam with a learning rate of $2\times10^{-5}$, weight decay of $0.05$, batch size 24, and a cosine learning-rate schedule with 15\% warmup steps and a cosine division factor of 10. We adopt the default image augmentation pipeline from the \href{https://github.com/paninski-lab/beast}{BEAST
 code repository}. Training is performed for 160 epochs for IBL and 400 for Cheese3D on a single NVIDIA A100 GPU and takes approximately 1 hour per session.


\subsection{TCN}
\label{app:tcn_training_details}
For the TCN encoder and decoder, we perform hyperparameter searches using Ray Tune to identify the best-performing configurations. Each TCN is trained using AdamW for 300 epochs with an MSE loss and a temporal-filter smoothing penalty of 0.5. We sample 30 random hyperparameter configurations, with learning rates drawn log-uniformly from $5\times10^{-5}$ to $2\times10^{-3}$. We apply a weight decay of $10^{-4}$ for model training. Model selection is performed using validation metrics ($R^2$ for decoding and bps for encoding). We then report results on the test set. Training and hyperparameter search are performed on 2 NVIDIA A40 or A100 GPUs, with the full search taking approximately 0.5 hours.

\end{document}

%% file: math_commands.tex
\usepackage{amsmath,amsfonts,bm}

\def\eqref#1{equation~\ref{#1}}

\def\1{\bm{1}}

\DeclareMathAlphabet{\mathsfit}{\encodingdefault}{\sfdefault}{m}{sl}
\SetMathAlphabet{\mathsfit}{bold}{\encodingdefault}{\sfdefault}{bx}{n}

